%% file: neurips_2026.tex
\documentclass{article}

\PassOptionsToPackage{numbers, compress}{natbib}

\usepackage[main, final]{neurips_2026}

\usepackage[utf8]{inputenc} 
\usepackage[T1]{fontenc}    
\usepackage{hyperref}       
\usepackage{url}            
\usepackage{booktabs}       
\usepackage{amsfonts}       
\usepackage{nicefrac}       
\usepackage{microtype}      
\usepackage{xcolor}         
\usepackage{comment}
\input{preamble}

\title{Gaussian Density Splatting Network}

\author{%
  Miao Shang \quad Yabin Wang \quad
  Xiaopeng Hong\thanks{Corresponding author.} \\
  Faculty of Computing \\
  Harbin Institute of Technology \\
  \texttt{miaos0522@gmail.com} \quad
  \texttt{wang-yabin@outlook.com} \quad
  \texttt{hongxiaopeng@ieee.org}
}

\begin{document}

\maketitle

\input{sec/0_abs.tex}
\input{sec/1_intro.tex}

\input{sec/2_related.tex}
\input{sec/3.tex}
\input{sec/4.tex}
\input{sec/5.tex}

{
    \bibliographystyle{unsrt}
    \bibliography{main}
}
\input{sec/Appendix}


\end{document}

%% file: preamble.tex
\usepackage{comment}
\usepackage{multirow}
\usepackage{enumitem}
\usepackage{tabularx}
\usepackage{framed}
\usepackage{amsthm}
\usepackage{amsmath}
\usepackage{amssymb}
\usepackage{xspace}
\usepackage{xcolor}
\usepackage{graphicx}
\usepackage{colortbl}

\usepackage{diagbox}
\usepackage{wrapfig}

\usepackage[ruled,vlined,linesnumbered]{algorithm2e}

\newcommand\Miao[1]{\textcolor{black}{#1}}
\newcommand\XP[1]{\textcolor{black}{#1}}

\usepackage[normalem]{ulem}
\usepackage{placeins} 
\newcommand{\CR}[1]{\textcolor{black}{#1}}

%% file: sec/0_abs.tex
\begin{abstract}

This paper proposes a novel crowd counting approach, the Gaussian Density Splatting Network (GDSNet). Unlike methods that rely on conventional, grid-based density maps and are sensitive to spatial resolution, GDSNet represents a crowd as a superposition of continuous 2D Gaussian primitives. Our approach is built upon two key contributions. 
First, a control-point-based fitting mechanism is introduced to structure the prediction of Gaussian parameters. A set of control points is adaptively allocated to define local regions, from which features are pooled to regress each primitive’s parameters. Second, we adapt a differentiable Gaussian Splatting framework to the counting task by parameterizing each primitive with geometric parameters and a scalar density mass. This formulation allows the network to be trained end-to-end via spatial matching of differentiably rendered density maps, naturally providing both local density supervision and global count optimization. Extensive evaluations on four standard benchmarks show GDSNet consistently outperforms the state of the art. The code is available at https://github.com/infinite0522/GDSNet-Gaussian-Density-Splatting-Network. 
\end{abstract}

%% file: sec/1_intro.tex
\section{Introduction}
\label{sec:intro}
\begin{figure}[t]
    \centering
    \vspace{-3mm}
    \includegraphics[width=0.9\linewidth]{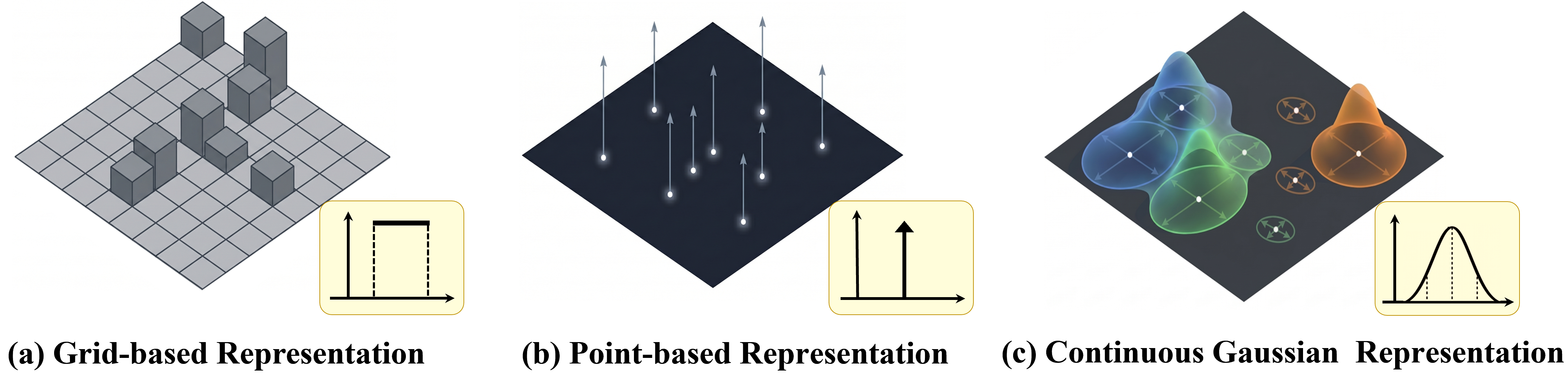}
    \vspace{-1mm}
    \caption{Comparison of grid-based, point-based, and our continuous Gaussian representation.}
    \vspace{-4.5mm}
    \label{fig:representation}
\end{figure}

{Crowd counting, the task of estimating the number and spatial distribution of people in an image, serves as a cornerstone of modern intelligent systems, with applications ranging from public safety monitoring to urban planning~\cite{zhang2016single, li2018csrnet, cao2018scale}.}
\XP{The core challenge of crowd counting lies in learning an expressive representation of the crowd density function, $f: \mathbb{R}^2 \to \mathbb{R}^+$, whose integral over the image domain yields the total count. However, existing paradigms for approximating this function suffer from fundamental representational limitations, as illustrated in Fig.~\ref{fig:representation}.}

{The dominant paradigm, grid-based density map regression, discretizes the domain and models the density function on a rigid grid~\cite{liu2019context, haniccvsteerer}. This approach is fundamentally constrained by a \emph{fixed} downsampling ratio of the network backbone, forcing the representation into a limited-resolution, piecewise-constant structure. This architectural rigidity not only introduces quantization artifacts but also struggles to model the continuous scale variations inherent in perspective-distorted scenes. A second paradigm, point-based detection, conceptualizes individuals as discrete points~\cite{song2021P2P, liang2022CLTR}. While continuous in their coordinates, these methods reduce people to Dirac delta-like impulses, ignoring the essential geometric properties of individuals and groups (such as spatial extent and orientation). This oversimplification, combined with a reliance on unstable bipartite matching for supervision, may lead to training instability and limited performance in dense regions~\cite{liu2023PET, chen2024APGCC}.}

{To overcome these limitations, we propose to represent the crowd density field as a superposition of continuous, learnable 2D Gaussian primitives. Each primitive is explicitly parameterized by its \emph{location}, anisotropic \emph{covariance} that captures scale and orientation, and scalar \emph{density mass} representing the local population count. This formulation elegantly unifies the localization-modeling problem (the geometric aspect) and the counting problem (the quantitative aspect) of crowd analysis.}

\XP{However, this powerful continuous representation introduces a significant optimization challenge, \emph{i.e.}, the learning problem becomes severely ill-posed. Unlike grid-based regression, where spatial support is predefined by pixels, learning a set of free-form Gaussians is prone to failure: First, without guidance, the network may waste modeling capacity in sparse areas while failing to represent high-density regions with sufficient fidelity. Second, the unconstrained optimization of covariance parameters can lead to degenerate solutions, where Gaussians collapse to points, stretch infinitely, or overlap chaotically, resulting in \emph{physically implausible representations}. Third, the Gaussian optimization procedure usually learns each Gaussian independently and ignores the spatial correlations between neighboring people, leading to a disjointed and locally incoherent density field.}

\XP{To address this ill-posed problem, we introduce a novel control-point guided approach that transforms unconstrained parameter regression into a structured geometric fitting problem. Our solution is built upon a hierarchy of components that directly counter the three challenges identified above. 
First, to satisfy the density-fidelity consistency, a structural complexity-guided control-point allocation strategy is introduced, which adaptively places points in structurally salient locations and further partitions the image into locally coherent regions. This directly prevents budget waste in homogeneous regions and perfectly aligns the primitive distribution with the non-uniform crowd density.
Second, to prevent physically implausible representations, these control points are organized via Delaunay triangulation to form a mesh of non-overlapping simplices.
\Miao{Each simplex provides a stable geometric prior for the Gaussian primitive and defines an explicit spatial boundary for its center location.} Finally, to ensure relational consistency, the final parameters of each Gaussian are predicted via a topology-aware feature aggregation scheme that considers the geometry and features of neighboring simplices. It substantially reduces optimization ambiguity and results in a stable Gaussian regression.}

\XP{To parameterize each Gaussian primitive based on a simplex formed by the control points, we design a Gaussian Density Splatting Network (GDSNet) to}
predict the final Gaussian parameters, including the location, covariance, and the density mass, by aggregating features from the local simplex geometry and the underlying image. The entire network is optimized end-to-end by differentiably rendering a density map from the learned primitives and matching it to a ground-truth map. Moreover, we introduce a count consistency loss and a shape regularization loss to enforce mass conservation and geometric stability.
\XP{Extensive experiments show that GDSNet sets a new state of the art on four challenging benchmarks and demonstrates superior robustness to variations in \Miao{internal feature resolution} at test time, validating the advantages of our continuous representation.}

Our main contributions are summarized below:
\begin{itemize}[noitemsep,topsep=0pt,parsep=2pt,partopsep=0pt,leftmargin=1em]

    \item \XP{We present a unified basis function-based crowd representation. Under this formulation, crowd density is modeled as a superposition of continuous, anisotropic 2D Gaussian primitives, overcoming the intrinsic limitations of discrete density maps and point-based representations.} 
    \item \XP{We propose a structural complexity-guided control-point allocation method to construct a triangulated topology, which provides structured spatial supports for organizing and parameterizing Gaussian primitives.}
    
    \item \XP{We develop an end-to-end differentiable Gaussian density splatting framework, where primitive parameters are predicted based on the control-point topology. This formulation jointly enables spatially dense supervision and accurate global count optimization, leading to consistent performance improvements across benchmarks.}
\end{itemize}

%% file: sec/2_related.tex
\vspace{-2mm}
\section{Related Work}
\vspace{-2mm}

\subsection{Crowd Density Representations}
\vspace{-1mm}

Existing crowd counting methods primarily rely on discrete representations, \XP{which can be further} categorized into \Miao{grid-based density} regression and point-based coordinate localization.

Traditional \textbf{map-based methods} project continuous crowds onto rigid two-dimensional grids, inevitably creating severe scale mismatches and blurry visual overlaps. To alleviate these defects, researchers introduced multiscale architectures~\cite{zhang2016single, li2018csrnet, liu2019context}, hierarchical density experts~\cite{du2023HMoDE}, and scale-aware transformers~\cite{lin2022boosting, lin2024gramformer}. \XP{Moreover}, intermediate targets evolved to prevent instance fusion through probability estimation~\cite{Cheng2021D2CNet}, inverse distance maps~\cite{liang2023FIDTM}, and regional integral supervision via Bayesian or spatial transport losses~\cite{ma2019bayesian, wang2020dm, ma2021learning, shang2026GST, lin2025pml}. \CR{Among these methods, GST~\cite{shang2026GST}and~\cite{shang2026untangling} fits appearance Gaussians offline to precompute a transport kernel for training supervision, while retaining a conventional network to regress the density map. Its Gaussians therefore serve as correspondence priors for supervision.}
Conversely, \textbf{point-based methods} treat crowd distributions as discrete Dirac impulses. Following the bipartite matching paradigm established by P2PNet~\cite{song2021P2P}, subsequent frameworks improved querying efficiency and stability via transformer processing~\cite{liang2022CLTR}, quadtree splitting~\cite{liu2023PET}, and localized constraints~\cite{chen2024APGCC, Lin_2025_p2r}. Predicting isolated proposals intrinsically collapses crowds into zero-dimensional coordinates, entirely ignoring continuous spatial extents. This point-to-point matching paradigm also becomes exceptionally fragile in ultra-dense scenes, resulting in severe training instability and unreliable gradient assignments.

Bridging these representational extremes, GDSNet elevates crowd counting to a continuous functional approximation using latent Gaussian bases. \CR{Unlike GST~\cite{shang2026GST}, GDSNet directly predicts mass-carrying Gaussians to represent and render the density field.} By explicitly modeling perspective geometry and mathematically decoupling global counts from spatial rendering grids, our framework establishes a matching-free and resolution-agnostic paradigm.

\subsection{Image Representation and Perception via Continuous Gaussians}

Gaussian splatting has significantly advanced visual representation by modeling scenes as continuous geometric mixtures in computer vision tasks such as compression~\cite{gaussianimage}, super-resolution~\cite{hu2024gaussiansr}, and inpainting~\cite{li20252dgsInpainting}.
Foundational optimization-based methods~\cite{gaussianimage, zhu2025large, zhang2025imagegs} achieve high-fidelity reconstruction and compression by leveraging multi-scale fitting and entropy-driven initialization. GaussianSR~\cite{hu2024gaussiansr} and GSASR~\cite{chen2025generalized} \XP{further} enable arbitrary-scale super-resolution via continuous field modeling. \XP{Nonetheless, the iterative optimization} remains computationally expensive.

To accelerate inference, feedforward \XP{2DGS} approaches have been developed to directly predict Gaussian parameters from visual features~\cite{hu2024gaussiansr, chen2025generalized,dong2025gaussiantoken}, \XP{which leverage neural networks (\emph{e.g.}, VAE, Transformer, UNet) to directly map image features to Gaussian fields.} There are increasing studies on incorporating priors to enhance the quality of representations~\cite{peng2025pixel2gauss, zeng2025instantGI, zhang2025gpstoken, li20252dgsInpainting}. Among them, Instant GaussianImage~\cite{zeng2025instantGI} represents observable RGB images through Gaussian-based interpolation.

These advances primarily target appearance reconstruction, leaving the formulation of task-oriented Gaussian fields for dense prediction less explored. GDSNet treats Gaussians as adaptive basis elements of an implicit continuous density field, with task-specific responses and spatial support. Through topology-aware primitive interaction and task-specific constraints, it adapts Gaussian representations to crowd density estimation.

%% file: sec/3.tex
\section{Method}
\label{sec:method}
\vspace{-0.5em}
\XP{We begin by revisiting density estimation from a representation perspective and reformulate it as continuous Gaussian field reconstruction in Sec.~\ref{subsec:representation}. We then introduce GDSNet to learn this representation, as shown in Fig.~\ref{fig:architecture} with its key components detailed in the following subsections.}

\vspace{-1mm}
\subsection{A Unified Mathematical Perspective on Crowd Representation}
\label{subsec:representation}
\vspace{-1mm}

Given an input image $I \in \mathbb{R}^{H \times W \times 3}$ and annotations $Y=\{y_m\}_{m=1}^{M}$, crowd counting aims to estimate a non-negative spatial density function $\hat{\mathcal{D}}(x)$ over the continuous domain $\Omega = \mathbb{R}^2$. From a functional approximation perspective, we unify existing paradigms 
as estimating a density field $\hat{\mathcal{D}}(x)$ via a linear combination of \textit{basis functions}:
\begin{equation}
    \hat{\mathcal{D}}(x) = \sum\nolimits_{i} \alpha_i \Psi_i(x; \xi_i),
    \label{eq:unified_basis}
\end{equation}
where $\Psi_i$ denotes the basis function whose spatial support is parameterized by $\xi_i$, $\alpha_i \ge 0$ is its magnitude. The global integral yields the total count $\hat{M} = \int_{\Omega} \hat{\mathcal{D}}(x)\,dx$. 

\textbf{Limitations of existing paradigms.} 
As summarized in Table~\ref{tab:paradigm_comparison}, existing methods are constrained by their suboptimal basis choices~\footnote{Detailed mathematical derivations and explanations of the formulas and notations in Table 1 are provided in \emph{Suppl.~\ref{sec:suppl_representation}}.}.
Grid-based methods instantiate $\Psi_i$ as a \textit{zero-order box basis with a rigid grid-aligned support}. The grid support is dictated by the network's downsampling stride, limiting flexibility for continuous density variations or sub-pixel congestions.
Point-based methods instantiate $\Psi_i$ as a \textit{Dirac delta basis with a zero-volume support}. The zero-volume support fails to explicitly model the spatial extent, local anisotropy, or continuous occupancy of crowds.

\textbf{Our solution.}
To achieve a representation that is both geometrically explicit and spatially continuous, we instantiate the basis $\Psi_i$ as an \textbf{\XP{anisotropic} 2D Gaussian} and form a continuous representation:
\begin{equation}
    \hat{\mathcal{D}}_{\mathrm{gds}}(x)
    =
    \sum\nolimits_{i}
    \rho_i \mathcal{N}(x;\mu_i,\Sigma_i).
    \label{eq:gsc_density}
\end{equation}
\XP{where $\mathcal{N}(x;\mu_i,\Sigma_i)$ is a normalized Gaussian distribution over $\mathbb{R}^2$.}
The magnitude $\alpha_i$ corresponds to a learnable density mass $\rho_i$. 
The count estimation is the summation of the density mass:
\begin{equation}
    \hat{M}
    =
    \int_{\Omega}\hat{\mathcal{D}}_{\mathrm{gds}}(x)\,dx
    =
    \int_{\Omega} \sum\nolimits_{i}
    \rho_i \mathcal{N}(x;\mu_i,\Sigma_i) \,dx
    =  \sum\nolimits_{i}
    \rho_i \int_{\Omega} \mathcal{N}(x;\mu_i,\Sigma_i) \,dx
    =
    \sum\nolimits_{i}\rho_i.
\end{equation}
\XP{The advantages are manifold.} First, the infinitely differentiable smooth basis \XP{provides a smooth and flexible function family capable of approximating continuous crowd density variations} to approximate the non-linear and smoothly decaying density distributions of real-world crowds under perspective variation and local occlusion. Second, count estimation is decoupled from the rasterization grid through mass summation, which reduces quantization errors.  \CR{Gaussian geometry nevertheless defines the spatial support of the density field and is jointly learned with mass through spatial supervision.} Third, the explicit primitive parameters make it possible to inject geometric and topological constraints into the learning process.

\begin{table}[t]
\centering
\caption{Comparison of crowd counting paradigms from a basis-function perspective.}
\label{tab:paradigm_comparison}
\resizebox{\textwidth}{!}{
\begin{tabular}{l|lllll}
\toprule
\textbf{Paradigm} & \textbf{Magnitude ($\alpha_i$)} & \textbf{Basis Function ($\Psi_i$)} & \textbf{Spatial Support ($\xi_i$)} & \textbf{State} & \textbf{Representation Formula $\hat{\mathcal{D}}(x)$} \\
\midrule
Grid-based   & Scalar density ($\hat{d}_u$) & Piecewise box ($\Pi_{\Omega_u}$)$ \ \ $    & Fixed grid cell ($s$)   & Discrete   & $\sum_{u \in \mathcal{G}_s} \hat{d}_u \Pi_{\Omega_u}(x)$ \\
Point-based & Binary indicator ($\mathbb{I}$)$ \ \ $  & Dirac delta ($\delta$)      & Zero-volume point ($\mu_i$) & Discrete   & $\sum_{i} \mathbb{I}(p_i>\eta)\delta(x-\mu_i)$ \\
\textbf{GDSNet} & \textbf{Density mass ($\rho_i$)} & \textbf{2D Gaussian  ($\mathcal{N}$)} & \textbf{Explicit geometry ($\mu_i, \Sigma_i$) $ \ \ $ } & \textbf{Continuous} $ \ \ $ & $\sum_{i} \rho_i \mathcal{N}(x;\mu_i,\Sigma_i)$ \\
\bottomrule
\end{tabular}
}
\end{table}

Despite its theoretical elegance, directly optimizing these Gaussian parameters is \XP{ill-posed due to non-identifiability and instability arising from unconstrained primitive locations and covariances.}
Without proper structural regulation, primitives easily collapse into singularities or migrate erratically. \XP{To impose structure on this otherwise unconstrained representation,} inspired by the construction of free-form curves approximation like Bézier and B-spline curves~\cite{bohm1984survey}, we propose to govern the continuous representation through a parametric modeling paradigm driven by structural control points. We reformulate the density estimation as a control-point-constrained parameterization problem, regulated by localized points that govern primitive centers and geometric priors.

\begin{figure}[t]
        \centering
        \includegraphics[width=1\linewidth]{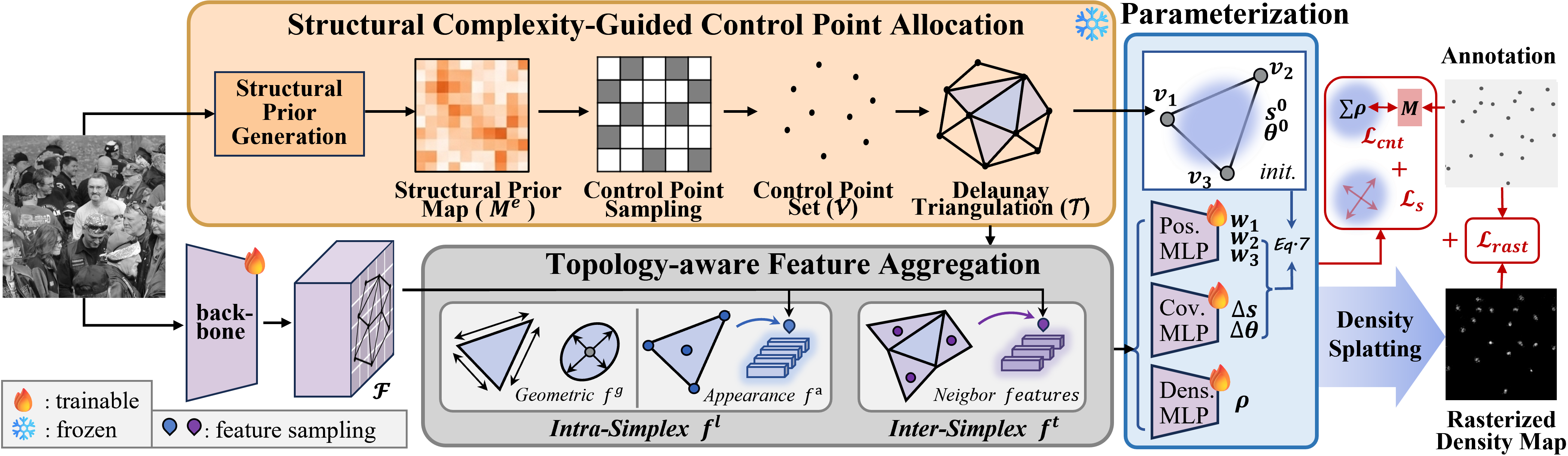}
        \caption{Pipeline of GDSNet.}
        \vspace{-1em}
        \label{fig:architecture}
\end{figure}

\Miao{This transforms free-form Gaussian estimation into a stable process of explicitly constrained parameterization. 
As in Fig.~\ref{fig:architecture}, we first perform structural complexity-guided control point allocation to construct a Delaunay triangulation $\mathcal T$ from the input image (Sec.~\ref{subsec:point_allocation}), while the counting backbone extracts a dense feature map. Driven by this spatial structure, we then perform topology-aware aggregation to fuse intra- and inter-features. The features are utilized to regress the Gaussian parameters and density mass, governed by a constrained parameterization mechanism (Sec.~\ref{subsec:parameterization}). Finally, the primitives undergo differentiable density rendering for end-to-end supervision (Sec.~\ref{subsec:optimization}).}

\subsection{Structural Complexity-Guided Control-Point Allocation}
\label{subsec:point_allocation}

The effectiveness of our parameterization fundamentally relies on the spatial distribution of control points. A naive uniform sampling allocates excessive modeling capacity to homogeneous regions while under-representing complex dense areas. 
To establish an adaptive spatial topology, we introduce a Structural Complexity-Guided Control-Point Allocation module. 

\textbf{Structural Guidance Sampling.}
Our design is motivated by the observation that local structural complexity reflects the underlying spatial layout of the scene. Regions exhibiting sharp appearance transitions, occlusion boundaries, or abrupt density variations typically present elevated structural complexity, indicating critical transitions in the underlying density field.

To capture critical scene transitions, we define a Structural Prior Map $\mathbf{M}^{e} \in [0,1]^{H \times W}$ that identifies regions of high structural complexity, such as sharp appearance transitions or occlusion boundaries. These regions require denser point allocation to ensure that the resulting interiors remain structurally coherent and ideally suited for compact Gaussian approximation. Thus, we generate $M^e$ by leveraging the off-the-shelf spatial structural estimator from the 2DGS framework~\cite{zeng2025instantGI}. 
This network was trained to predict the optimal locations for placing Gaussians to minimize rendering error, which naturally correlates with high-frequency details, edges, and structural complexity.
We provide a detailed discussion on alternative $M^e$ in Sec.~\ref{exp: map_generation}.

To convert the structural prior $\mathbf{M}^{e}$ into a discrete point set while maintaining explicit control over the representational granularity, we introduce a stride-controlled sampling mechanism, aiming to regulate both the total number of Gaussian primitives and their average spatial bounds across the continuous domain.
We first apply non-overlapping max pooling with kernel size and stride $s$ to $\mathbf{M}^{e}$, producing a coarse structural response map $\bar{\mathbf{M}}^{e}$. A Floyd-Steinberg dithering operator~\cite{franchini2019stochastic} is then applied to yield binary activations $z_{i,j}=\mathrm{Dither}(\bar{\mathbf{M}}^{e}_{i,j})\in \{0,1\}$. The resulting control-point set is defined as:
\begin{equation}
\mathcal{V} = \left\{
(is+\tfrac{s}{2}, js+\tfrac{s}{2})
\mid z_{i,j}=1
\right\}.
\end{equation}
This strategy guarantees that the control points are adaptively distributed based on local structure, while their total quantity $|\mathcal{V}|$ is strictly bounded by $\lceil H/s \rceil \times \lceil W/s \rceil$. 

The stride $s$ explicitly dictates the spatial resolution of the continuous Gaussian field. Because this continuous paradigm fundamentally decouples density modeling from the fixed pixel grid, modulating $s$ simply adjusts the representation's granularity.  This intrinsic flexibility endows the framework with a strong robustness against varying representation granularities, thereby allowing GDSNet to perform stable, arbitrary test-time scaling without retraining, as validated experimentally in Sec.~\ref{subsec:sensitivity}.

\textbf{Delaunay simplex construction.}
Given the sampled control-point set $\mathcal V$, we \XP{perform planar Delaunay triangulation~\cite{DelaunayTri} to 
obtain a set of simplices}
$\mathcal T
=
\mathrm{Del}(\mathcal V)
=
\{\tau_i\}_{i=1}^{N},
$
where each simplex $\tau_i = \text{conv}(\{v_{i,1}, v_{i,2}, v_{i,3}\})$ is \XP{a convex hull} defined by a triplet of vertices from $\mathcal V$.

This process yields triangular support domains that align with the structural boundaries identified in the sampling stage. The choice of 2D simplices over arbitrary polygons is motivated by three critical advantages for Gaussian parameterization.
First, simplices \emph{enable stable and unique geometric regression through barycentric coordinates~\cite{cabral1999reflection}}, ensuring smooth gradient flow by restricting primitive centers within a convex feasible region.   
Second, the Delaunay \emph{empty circumcircle} property~\cite{DelaunayTri} maximizes the minimum internal angle of the mesh, preventing spatially elongated primitives that would violate local density assumptions. Finally, by Euler's formula, $|\mathcal{V}|$ control points yield at most $2|\mathcal{V}| \!-\!5$ faces~\cite{seidel1995upper}, \emph{bounding the final number of generated Gaussians to $\mathcal{O}(|\mathcal{V}|)$} while guaranteeing a completely hole-free, continuous representation of the crowd.
By ensuring that each simplex encloses a region of relative informational stability, this geometric partitioning provides a locally coherent substrate for compact Gaussian approximation.

\subsection{Topology-Aware Feature Aggregation and Constrained Parameterization}
\label{subsec:parameterization}

Given the control points $\mathcal{V}$ and the simplex set $\mathcal{T}$, the subsequent task is to determine the specific geometric configuration and density mass for each Gaussian primitive. We design a topology-aware feature aggregation module that operates over a topological graph to extract hierarchical features, followed by a formal geometric parameterization that enforces spatial constraints. To capture latent structural dependencies among primitives, we formalize the relational structure of the support domains as a topological graph $\mathcal{G} = (\mathcal{T}, \mathcal{E})$. The connectivity in $\mathcal{G}$ is governed by the shared edges \XP{$\mathcal{E}$} of the Delaunay triangulation, \XP{with} the neighborhood for each simplex $\tau_i$ $\mathcal{B}(i) = \{ \tau_j \mid \tau_i \cap \tau_j \in \mathcal{E} \}$ represents the set of adjacent simplices sharing a common edge.

\textbf{Topology-aware Feature Aggregation.}
We extract and aggregate features for each primitive over the simplicial graph $\mathcal{G}$ to encode both local and global contexts. Specifically, the feature $f_i$ for the $i$-th Gaussian primitive consists of an intra- and inter-simplex representations, $f_i^l$ and $f_i^t$, respectively.

\textit{Intra-simplex feature representation.} 
The intra-simplex representation $f_i^l$ fuses a geometric representation $f_i^g$ with an appearance feature $f_i^a$. Specifically, $f_i^g$ is explicitly constructed using the normalized coordinates of the triangle vertices alongside the parameters of the base ellipse $e_i$ fitted to the simplex~\cite{fitzgibbon1996buyer}. The appearance feature $f_i^a$ is obtained by querying the dense feature map $\mathcal{F} \in \mathbb{R}^{C \times H' \times W'}$ extracted by the counting backbone. To capture the local appearance context across this support region, we sample $\mathcal{F}$ at the triangle barycenter $c_i$ and its three vertices $v_{i,j}$:
\begin{equation}
    f^a_i = \text{MLP}_\textit{app}\left( \mathcal{F}(c_i) \oplus \mathcal{F}(v_{i,1}) \oplus \mathcal{F}(v_{i,2}) \oplus \mathcal{F}(v_{i,3})\right),
\label{eq:app_feature}
\end{equation}
where $\text{MLP}_\textit{app}$ is designed to aggregate these queries.

\textit{Inter-simplex topological aggregation.} 
Due to the spatial correlation inherent in crowd distributions, adjacent local regions commonly exhibit similar perspectives, scales, and density patterns. Treating each triangular receptive field independently would neglect this relational structure, leading to incoherent parameter predictions. Therefore, we perform topological context aggregation over the graph $\mathcal{G}$ to ensure consistency across neighboring primitives. For each triangle $\tau_i$, we collect its one-hop adjacent neighbors $\mathcal{B}(i)$ and aggregate their graph-induced contextual cues to form $f_i^t$, which is then concatenated with the intra-simplex representation:
\begin{equation}
    f_i = f_i^l \oplus f_i^t, \quad \text{where} \quad f_i^t =  \bigoplus\nolimits_{b \in \mathcal{B}(i)} \text{MLP}_\textit{nbr}(c_b \oplus \mathcal{F}(c_b)).
    \label{eq:fi}
\end{equation}
$c_b$ denotes the barycenter of a neighboring triangle $b$, and $\text{MLP}_\textit{nbr}$ compresses the neighbor's feature channels. 
\XP{Eq.~\ref{eq:fi} allows the subsequent parameterization to incorporate local perspective and density variations, thereby reducing geometric and spatial inconsistencies across neighboring regions.}

\textbf{Control-Point-Constrained Parameterization. }
To facilitate stable Gaussian learning, we obtain the final state of each primitive $\mathcal{G}_i = \{\mu_i, \mathbf{s}_i, \theta_i, \rho_i\}$ through a constrained parameterization governed by the underlying support simplex. Rather than directly regressing these values in an unconstrained domain, we formulate the final parameters as regulated adjustments relative to a base prior.

Specifically, the reference ellipse $e_i$ fitted over the simplex $\tau_i$ establishes a plausible base prior. Taking the aggregated feature representation $f_i$ as input, parallel MLP heads predict a set of relative adjustments alongside the non-negative density mass $\rho_i$. These geometric adjustments include the barycentric weights $\mathbf{w}_i = [w_{i,1}, w_{i,2}, w_{i,3}]$, the scale offsets $\Delta \mathbf{s}_i$, and the rotation offset $\Delta \theta_i$. 

Using the network predictions together with the initial rotation $\theta_i^{\mathrm{0}}$ and base aspect ratio $r_i^{\mathrm{0}}\!=\! s^{\mathrm{0}}_{i,x}/(s^{\mathrm{0}}_{i,y}\!+\!\epsilon)$ derived from $e_i$, the final geometric parameters of the Gaussian primitive are:
\begin{equation}
\mu_i = \sum\nolimits_{j=1}^3 w_{i,j} v_{i,j}, \quad \mathbf{s}_i = \Delta \mathbf{s}_i \cdot \begin{bmatrix} r_i^{\mathrm{0}} , 1 \end{bmatrix}^\mathsf{T}, \quad \theta_i = \theta_i^{\mathrm{0}} + \Delta \theta_i.
\label{eq:mu_sigma}
\end{equation}
Crucially, the optimization of the Gaussian center $\mu_i$ is explicitly anchored by the learned barycentric weights \cite{cabral1999reflection}. This restricts the center to a localized feasible region bounded by the convex hull of its associated control points $\mathcal{R}_i = \{x \mid x = \sum\nolimits_{j=1}^3 w_{i,j} v_{i,j}, \sum\nolimits_{j=1}^3 w_{i,j}=1, w_{i,j} \ge 0\}$. \XP{This formulation establishes a rigid spatial boundary that prevents erratic primitive migration.}

\subsection{Continuous Density Rendering and Optimization}
\label{subsec:optimization}
\vspace{-0.5em}
\XP{Building upon these parameterized Gaussian primitives, GDSNet adapts the differentiable Gaussian splatting framework from traditional RGB color synthesis to crowd density estimation. Specifically, the continuous density field is formulated by aggregating all Gaussian primitives, each weighted by its corresponding scalar density mass $\rho_i$, as defined in Eq.~\ref{eq:gsc_density}.}

For spatial density alignment, we rasterize the density map and apply Bayesian loss~\cite{ma2019bayesian}:
\begin{equation}
    \mathcal{L}_{\mathrm{rast}} = \sum\nolimits_{m=1}^M \left| 1 - \int_{\Omega} \mathbf{P}_m(x) \hat{\mathcal{D}}(x) \,dx \right| + \left| 0 - \int_{\Omega} \mathbf{P}_{bk}(x) \hat{\mathcal{D}}(x) \,dx \right|,
\label{eq:rast_loss}
\end{equation}
$\mathbf{P}_{m}$ and $\mathbf{P}_{bk}$ are the foreground and background posterior probability maps as in~\cite{ma2019bayesian}. 

To further enforce global counting consistency, we introduce a count conservation loss
\begin{equation}
\mathcal L_{\mathrm{cnt}}
=
\left|
M-\sum\nolimits_{i=1}^{N}\rho_i
\right|.
\label{eq:cnt_loss}
\end{equation}

To discourage excessively elongated Gaussian shapes, we apply a shape regularization term:
\begin{equation}
\mathcal{L}_{\text{s}}
= \max_i \left[ {s_i^{\max}}/({s_i^{\min}+\epsilon})-\gamma \right]_+,
\label{eq:shape_loss}
\end{equation}

where $s_i^{\max}$ and $s_i^{\min}$ denote the major and minor axis scales of the $i$-th Gaussian, respectively, and $\gamma$ controls the maximum allowable aspect ratio. The overall loss function is:
\begin{equation}
\label{eq:overall}
\mathcal L
=
\mathcal L_{\mathrm{rast}}
+
\lambda_1 \mathcal L_{\mathrm{cnt}}
+
\lambda_2 \mathcal L_{\mathrm{s}}.
\end{equation}
\XP{Eq.~\ref{eq:overall} promotes density reconstruction, count preservation, and Gaussian geometry fitting jointly.}

%% file: sec/4.tex
\vspace{-0.5em}
\section{Experiment Results and Discussion}
\vspace{-0.5em}
\label{sec:exp}
\label{subsec:setup}

We evaluate GDSNet on four \XP{standard} crowd benchmarks: ShanghaiTech A and B~\cite{shanghai}, JHU-Crowd++~\cite{sindagi2020jhu}, UCF-QNRF~\cite{ucfqnrf}, and NWPU-Crowd~\cite{nwpu}, using MAE and MSE as the performance metrics.
Following previous competitive methods~\cite{chen2024APGCC,lin2025pml}, we utilize the VGG-19 structure as our feature extractor. 
Default sampling stride $s$ is $10$. $\lambda_1\!=\!\lambda_2=1$. \XP{AdamW optimizer is used}, with a learning rate of $1\times10^{-5}$, a weight decay of $1\times10^{-4}$, and a batch size of 1. All experiments are conducted on NVIDIA RTX 4090 GPUs. Details and the implementation algorithm are in \emph{Suppl.~\ref{sec:supp_implementation}}.

\subsection{Evaluation Results}
\vspace{-0.5mm}
\label{subsec:sota}
We compare GDSNet with representative map-based and point-based methods in Table~\ref{tab:counting comparison}. GDSNet achieves a state-of-the-art performance across \emph{all} benchmarks. 
Specifically, GDSNet achieves a significant MAE reduction over the latest point-based method APGCC~\cite{chen2024APGCC} on JHU++, UCF-QNRF, and NWPU. This is attributed to our continuous Gaussian representation. \CR{On ShanghaiTech Part B, GDSNet achieves an MAE of 5.4 and an MSE of 8.0, demonstrating its effectiveness in relatively sparse crowd scenes.} By rasterizing Gaussian primitives into a structured density map, we leverage advanced dense supervision from map-based methods, circumventing the bipartite matching instabilities prevalent in ultra-dense regions. \XP{Moreover, GDSNet clearly outperforms recent map-based methods like PML~\cite{lin2025pml} and GST~\cite{shang2026GST}}, demonstrating the advantages of using continuous basis formulation over rigid grid-aligned scalar fields.

\begin{figure}[thbp!]
    \centering
    \includegraphics[width=1\linewidth]{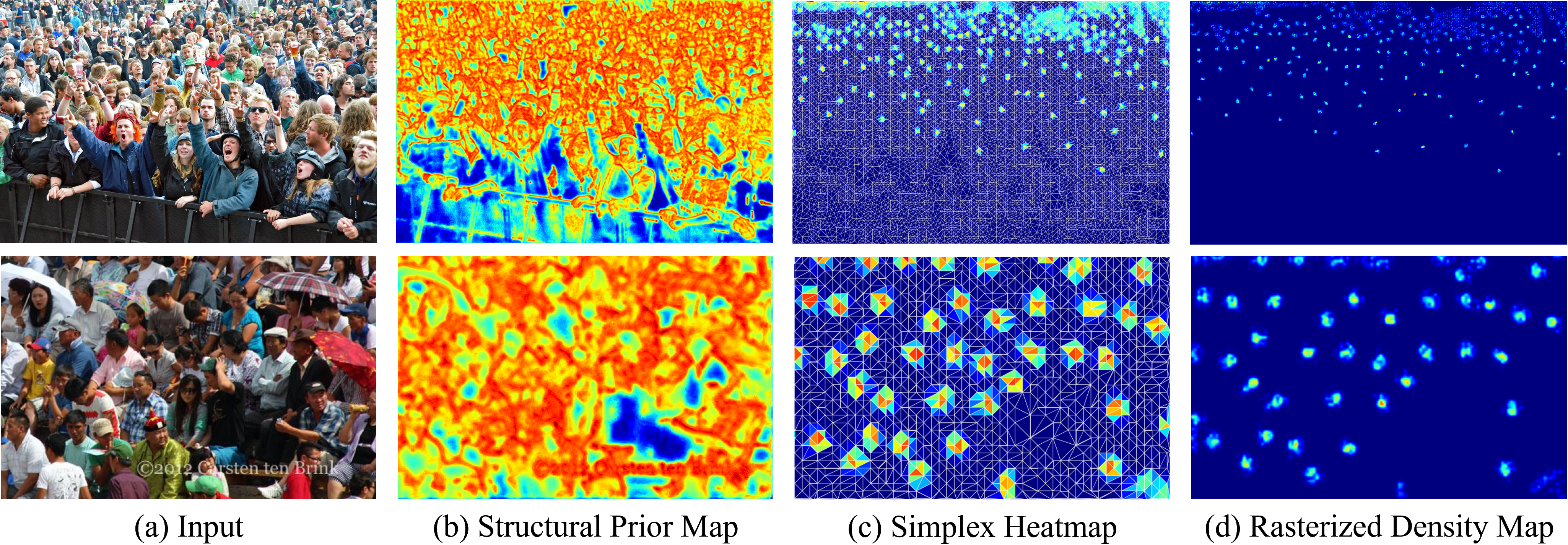}
    \vspace{-0.5mm}
    \caption{Visualization of the structural prior, simplex heatmap, and rasterized density map. The simplex heatmap is obtained by assigning each Delaunay simplex the mass predicted by its associated Gaussian primitive, revealing how responses are spatially distributed over the triangulated support.}
    \label{fig:vis}
    \vspace{-1em}
\end{figure}

\textbf{Visualization.}
Fig.~\ref{fig:vis} visualizes the intermediate representations of GDSNet across three crowd scenes. These examples exhibit \XP{diverse resolutions and} significant inter-image and intra-image scale variations, ranging from dense congregations to sparser scenes with varying perspective distortions. The structural prior maps in Fig.~\ref{fig:vis}~(b) identify structural boundaries and high transitions to guide the adaptive allocation of control points. In Fig.~\ref{fig:vis}~(c), \XP{the resulting Delaunay triangulation demonstrates strong adaptability across diverse scenes. Specifically, the total number and spatial distribution of the triangular simplices naturally align with the underlying crowd density.}
Crucially, the triangle mass heatmap reveals a smooth density transition across adjacent simplices, which confirms that our neighbor context aggregation effectively maintains topological continuity and captures structural dependencies between neighboring primitives. Finally, the rasterized density maps in Fig.~\ref{fig:vis}~(d) showcase the continuous density field formed by the superposition of Gaussian primitives, \XP{which} accurately capture the local geometry and density of the scene.

\begin{table*}[thbp!]
\setlength{\tabcolsep}{1.6mm}
\footnotesize
\centering
 \begin{tabular}{c|c|cc|cc|cc|cc|cc}
 \toprule
 \multicolumn{1}{c|}{Dataset} &\multirow{2}{*}{Venue \& Year} & \multicolumn{2}{c|}{ShTech A} & \multicolumn{2}{c|}{\CR{ShTech B}} & \multicolumn{2}{c|}{JHU-Crowd++} & \multicolumn{2}{c|}{UCF-QNRF} & \multicolumn{2}{c}{NWPU}\\ 
 \multicolumn{1}{c|}{Method} & &
  \multicolumn{1}{c}{MAE} & \multicolumn{1}{c|}{MSE} &
  \multicolumn{1}{c}{\CR{MAE}} & \multicolumn{1}{c|}{\CR{MSE}} &
  \multicolumn{1}{c}{MAE} & \multicolumn{1}{c|}{MSE} &
  \multicolumn{1}{c}{MAE} & \multicolumn{1}{c|}{MSE} &
  \multicolumn{1}{c}{MAE} & \multicolumn{1}{c}{MSE}\\
 \hline
 BL~\cite{ma2019bayesian} & ICCV 2019 & 62.8 & 101.8 & \CR{7.7} & \CR{12.7} & 75.0 & 299.9 & 88.7 & 154.8 & 105.4 & 454.2 \\
 DMCount~\cite{wang2020dm} & NeurIPS 2020 & 59.7 & 95.7 & \CR{7.4} & \CR{11.8} & 61.6 & 256.1 & 85.6 & 148.3 & 88.4 & 388.6 \\
 UOT~\cite{ma2021learning} & AAAI 2021 & 58.1 & 95.9 & \CR{6.5} & \CR{10.2} & 60.5 & 252.7 & 83.3 & 142.3 & 87.8 & 387.5 \\
 GL~\cite{wan2021generalized} & CVPR 2021 & 61.3 & 95.4 & \CR{7.3} & \CR{11.7} & 59.5 & 259.5 & 84.3 & 147.5 & 79.3 & 346.1 \\
 P2PNet~\cite{song2021P2P} & ICCV 2021 & 52.7 & 85.1 & \CR{6.3} & \CR{9.9} & -- & -- & 85.3 & 154.5 & 72.6 & 331.6 \\
 MAN~\cite{lin2022boosting} & CVPR 2022 & 56.8 & 90.3 & \CR{--} & \CR{--} & 53.7 & \textbf{209.9} & 77.3 & 131.5 & 76.5 & 323.0 \\
 CLTR~\cite{liang2022CLTR} & ECCV 2022 & 56.9 & 95.2 & \CR{6.5} & \CR{10.6} & 59.5 & 240.6 & 85.8 & 141.3 & 74.3 & 333.8 \\
 ChfL~\cite{shu2022chfl} & CVPR 2022 & 57.5 & 94.3 & \CR{6.9} & \CR{11.0} & 57.0 & 235.7 & 80.3 & 137.6 & 76.8 & 343.0 \\
 PET~\cite{liu2023PET} & ICCV 2023 & 49.3 & 78.8 & \CR{6.2} & \CR{9.7} & 58.5 & 238.0 & 79.5 & 144.3 & 74.4 & 328.5 \\
 APGCC~\cite{chen2024APGCC} & ECCV 2024 & 48.8 & 76.7 & \CR{5.6} & \CR{8.7} & 54.3 & 225.9 & 80.1 & 136.6 & 71.7 & \textbf{284.4} \\
 Gramformer~\cite{lin2024gramformer} & AAAI 2024 & 54.7 & 87.1 & \CR{--} & \CR{--} & 53.1 & 228.1 & 76.7 & \textbf{129.5} & 72.5 & 316.4 \\
 PML~\cite{lin2025pml} & ICLR 2025 & 55.5 & 89.0 & \CR{6.0} & \CR{9.3} & 57.4 & 227.4 & 76.6 & 132.2 & 73.6 & 338.6 \\
\CR{P2R~\cite{Lin_2025_p2r}} & \CR{CVPR 2025} & \CR{51.0} & \CR{79.7} & \CR{6.2} & \CR{9.8} & \CR{58.8} & \CR{253.1} & \CR{83.3} & \CR{138.1} & \CR{--} & \CR{--}\\
 GST~\cite{shang2026GST} & AAAI 2026 & -- & -- & \CR{--} & \CR{--} & 53.9 & \underline{225.4} & 80.7 & 131.1 & 74.4 & 306.2 \\
 \rowcolor{gray!15}
GDSNet & -- & \textbf{48.8} & \textbf{76.5} & \CR{\textbf{5.4}} & \CR{\textbf{8.0}} & \textbf{52.8} & \shortstack{227.6} & \textbf{75.9} & \underline{130.5} & \textbf{68.7} & \underline{294.2}\\
 \bottomrule
 \end{tabular}
\caption{Counting performance on ShTech A \CR{and B}, JHU-Crowd++, UCF-QNRF, and NWPU.}
\label{tab:counting comparison}
\end{table*}

\vspace{-2mm}
\subsection{Ablation Study}
\vspace{-1mm}
\label{subsec:ablation}

\textbf{Contribution of Components.}
\label{component_ab}
We investigate the contribution of the structural complexity-guided Control Point Allocation~(CPA), Simplex-Constrained Parameterization~(SCP), and Topology-aware Feature Aggregation~(TFA) in GDSNet on the UCF-QNRF dataset in Table~\ref{tab:ablation_modules}. 
Removing the topological context aggregation (\emph{Row 2}) leads to a performance drop, confirming that explicitly modeling local neighborhood interactions is important for preserving relational consistency among adjacent Gaussian primitives and producing spatially coherent density estimation. Moreover, the substantial degradation in (\emph{Rows 3\&4}) highlights the critical role of our simplex-constrained geometric formulation. 
Without the explicit geometric boundaries imposed by local simplices, Gaussian parameter learning becomes severely under-constrained, allowing the regression head to fit training samples through arbitrary scale and shape configurations that lack structural regularity. Although these unconstrained variants can achieve satisfactory fitting on the training set, they exhibit significantly weaker generalization on the test set, indicating that simplex support serves as an effective geometric prior that regularizes parameter optimization toward stable and transferable Gaussian representations. Furthermore, \emph{Row 3} outperforms \emph{Row 4}, further suggesting that structural complexity-guided sampling provides more informative primitive allocation than uniform grid sampling. \XP{Details are provided in \emph{Suppl.~\ref{sec:supp_module_ab}}.}

\begin{table*}[ht]
\vspace{-1em}
\centering
\footnotesize
\setlength{\tabcolsep}{2pt}
\begin{minipage}[t]{0.40\textwidth}
\vspace{0pt}
\centering
\setlength{\tabcolsep}{6pt} 
\caption{Ablation of key components.}
\label{tab:ablation_modules}
\begin{tabular}{ccc|cc}
\toprule
CPA & SCP & TFA & MAE & MSE \\
\midrule
\checkmark & \checkmark & \checkmark & \textbf{75.86} & \textbf{130.52} \\  
\checkmark & \checkmark & & 77.89 & 132.59 \\
\checkmark & & & 82.94 & 135.62 \\
& & & 89.46 & 140.85 \\    
\bottomrule
\end{tabular}
\end{minipage}
\hfill
\begin{minipage}[t]{0.28\textwidth}
\vspace{0pt}
\centering
\caption{Ablation of losses.}
\label{tab:ablation_loss}
\vspace{2pt}
\begin{tabular}{ccc|cc}
\toprule
$\mathcal{L}_{\text{rast}}$ & $\mathcal{L}_{\text{cnt}}$ & $\mathcal{L}_{\text{geo}}$ & MAE & MSE \\
\midrule
\CR{--} & \CR{$\checkmark$} & \CR{$\checkmark$} & \CR{94.87} & \CR{154.24} \\
\checkmark & & & 77.54 & 133.17 \\
\checkmark & \checkmark & & 76.87 & 131.58 \\
\checkmark & & \checkmark & 76.20 & 130.74 \\
\checkmark & \checkmark & \checkmark & \textbf{75.86} & \textbf{130.52} \\
\bottomrule
\end{tabular}
\end{minipage}
\hfill
\begin{minipage}[t]{0.30\textwidth}
\vspace{0pt}
\centering
\setlength{\tabcolsep}{2pt} 
\caption{CPA Strategies.}
\label{tab:ablation_init}
\begin{tabular}{l|cc}
\toprule
Strategy & MAE & MSE \\
\midrule
Uniform & 81.65 & 133.97 \\
\XP{Sobel} & 78.44 & 131.58 \\
\XP{Structural} & \textbf{75.86} & \textbf{130.52} \\
\bottomrule
\end{tabular}
\end{minipage}
\vspace{-0.5em}
\end{table*}

\CR{\textbf{Gaussian Geometry Ablation.}
We investigate the role of Gaussian geometry on UCF-QNRF in Table~\ref{tab:cr_geometry}. Triangle mass regression uniformly distributes the predicted mass over each triangle. Fixed Gaussians use the center, scale, and orientation of the reference ellipse without learning geometric adjustments. Isotropic Gaussians restrict spatial support to a single scale shared across both axes, whereas the full model learns anisotropic scales and orientation with simplex-constrained centers. All variants share the same primitive allocation, feature aggregation, mass prediction head, and rasterization supervision.
As shown in Table~\ref{tab:cr_geometry}, compared with triangle mass regression, Gaussian bases provide smooth density responses rather than piecewise-constant distributions. Learnable geometry allows spatial support to adapt to local crowd structure instead of remaining fixed to the initial triangulation. Anisotropic geometry further captures directional variations and elongated density patterns that isotropic support cannot represent.
Implementation details are in \emph{Suppl.~\ref{sec:cr_geometry}}.}

\begin{table}[htbp]
\centering
\footnotesize
\caption{Gaussian geometry ablation on UCF-QNRF.}
\label{tab:cr_geometry}
\begin{tabular}{l|cc}
\toprule
Configuration & MAE & MSE\\
\midrule
Triangle mass regression & 82.21 & 133.85\\
Fixed Gaussian & 81.42 & 133.24\\
Isotropic Gaussian & 78.79 & 132.86\\
Full anisotropic Gaussian & \textbf{75.86} & \textbf{130.52}\\
\bottomrule
\end{tabular}
\end{table}

\textbf{Loss Function Ablation.}
The optimization of GDSNet is guided by a joint objective that addresses the unique characteristics of our Gaussian representation. \XP{Table~\ref{tab:ablation_loss} shows that every loss term contributes by providing constraints from different perspectives.}
The rasterization loss serves as the primary driver by providing supervision on the rendered density field to ensure spatial fidelity. Furthermore, the count loss further refines the estimation by supervising the total density masses of Gaussian primitives. 
More importantly, geometric regularization plays a vital role in stabilizing primitive shapes by penalizing extreme anisotropies. 
The synergy of these objectives allows GDSNet to achieve the best accuracy by balancing spatial fidelity, counting precision, and geometric stability.

\textbf{Impact of Control Point Allocation Strategies.}
\label{exp: map_generation}
We evaluate \XP{three control point allocation strategies: \emph{baseline uniform sampling}, \emph{sampling according to the Sobel filtering responses~\cite{zhang2025imagegs}}, and our \emph{structural complexity-guided sampling}.}
Details and more prior map visualization are in \emph{Suppl.~\ref{sec:supp_CPA_ab}}.

As in Table~\ref{tab:ablation_init}, 
\emph{uniform sampling} yields the highest errors as it allocates control points indiscriminately across the image domain. The support simplex obtained from these control points becomes homogeneous because they have the same geometric characteristics. Compared with ours, \emph{sampling by Sobel filtering responses} is less effective because it is overly sensitive to sharp boundaries and fine textures. This over-sensitivity results in an excessive allocation of primitives in semantically irrelevant regions, which \XP{hurts the counting task.}

\begin{wrapfigure}[17]{r}{0.50\textwidth}
\vspace{-8pt}
\centering
\includegraphics[width=\linewidth]{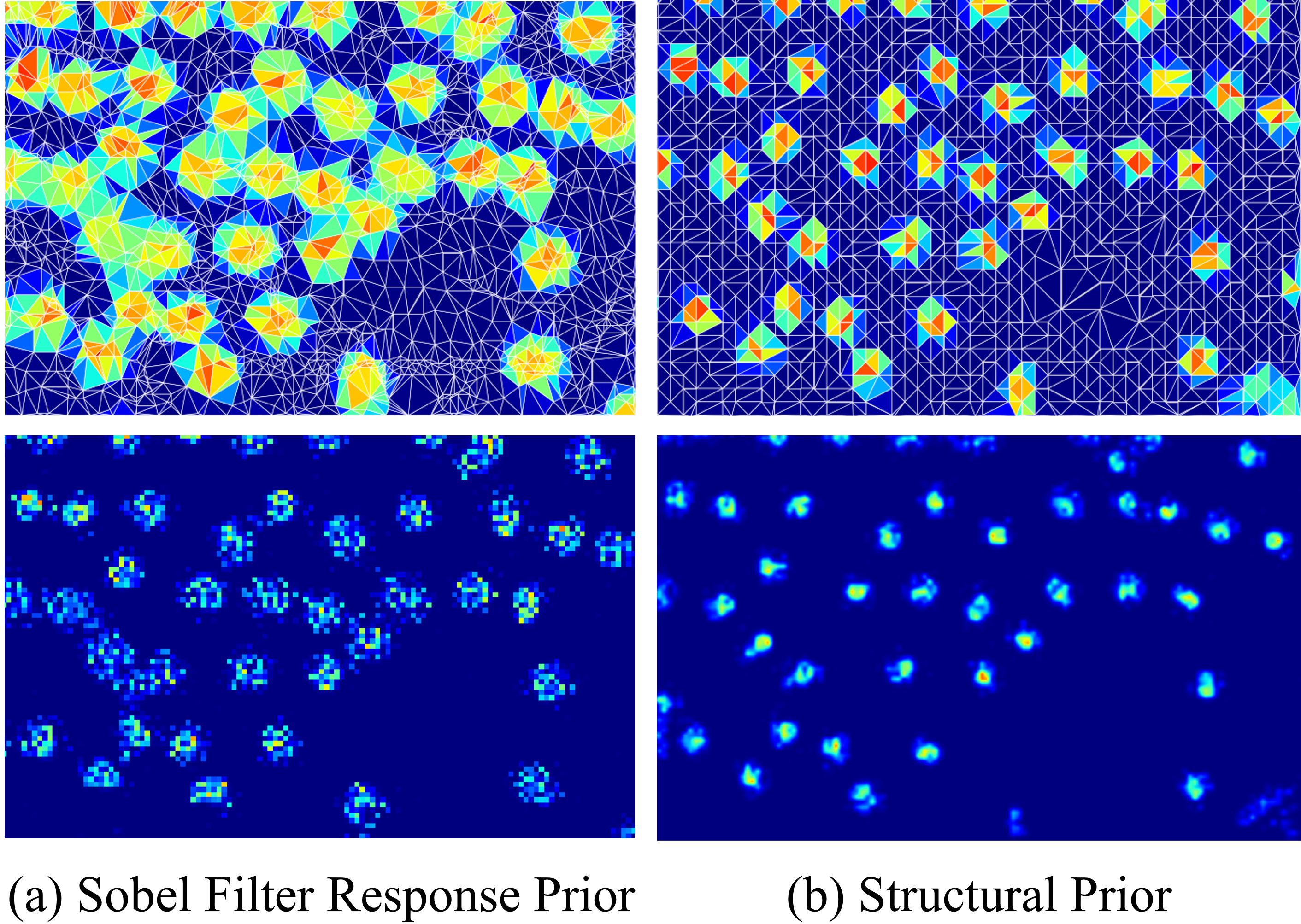}
\vspace{-1.5em}
\caption{Simplex heatmap (top) and rasterized density map (bottom) of different CPA priors.}
\label{fig:cpa_vis}
\end{wrapfigure}
\XP{Our \emph{structural complexity guided sampling} performs best.}
Taking the 3rd sample from Fig.~\ref{fig:vis} as an example, Fig.~\ref{fig:cpa_vis} \XP{further visualizes} the simplex heatmap and rasterized density maps.
Under the same sampling stride, \emph{sampling by Sobel filtering responses} activates a larger number of broad-response simplices around each head instance. To satisfy the unit-mass constraint, optimization compensates by aggressively shrinking individual Gaussian scales, resulting in fragmented and pulse-like density responses that undermine local spatial continuity. In contrast, our \emph{structural complexity guided sampling} produces more compact and spatially focused simplex activations, enabling Gaussian primitives to maintain appropriate support extent and generate smoother, more coherent density reconstruction. \XP{Enhanced alignment} between simplex geometry and crowd structure ultimately leads to more stable Gaussian parameterization and improved counting accuracy.

\subsection{Discussion on Sampling Stride Sensitivity}
\label{subsec:sensitivity}
\begin{figure}[htbp!]
    \centering
    \includegraphics[width=0.9\linewidth]{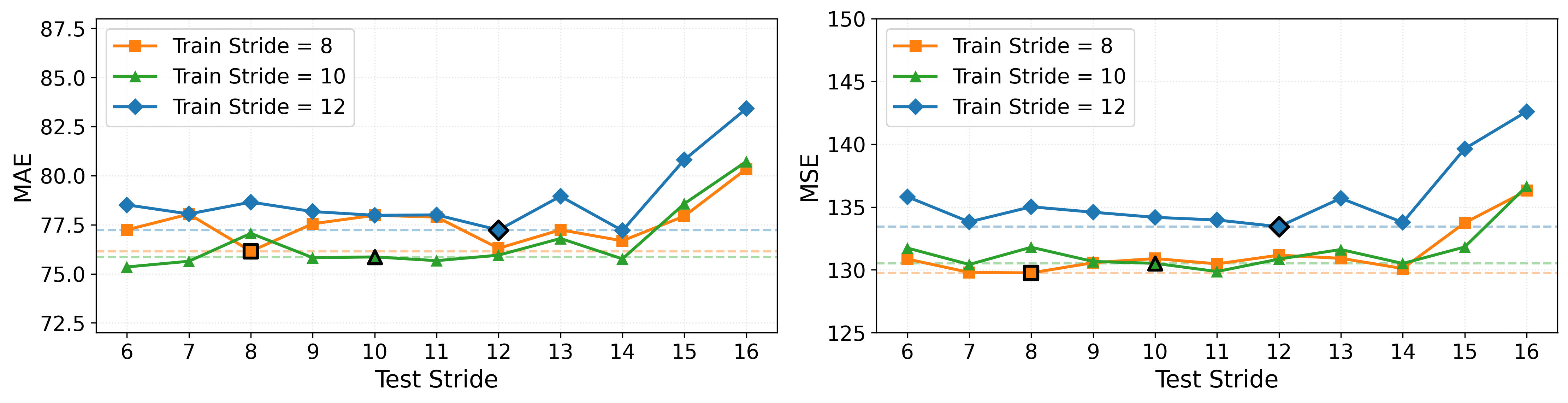}
    \caption{Performance \emph{w.r.t.} train and test sampling strides.}
    \label{fig:stride}
\end{figure}

\CR{\textbf{Robustness.}} To evaluate robustness under varying sampling resolutions, we conduct a train-test stride cross-ablation on UCF-QNRF by training GDSNet with strides 8, 10, and 12, and evaluating each across test strides from 6 to 16 without finetuning.
As shown in Fig.~\ref{fig:stride}, all configurations maintain MAE below 80 for test strides \XP{less} than 15, demonstrating strong robustness across a broad sampling range. This verifies that the continuous Gaussian representation effectively reduces dependence on a fixed discrete sampling resolution. Performance drops consistently at test strides 15 and 16 for all settings, revealing an intrinsic undersampling limit rather than train-test mismatch. At such sparse resolutions, Gaussian primitives become excessively coarse, leading to merged local structures and over-smoothed density reconstruction.
Training with strides 8 and 10 achieves the strongest overall performance, with stride 10 performing best in most cases. Notably, it even surpasses its matched-stride result when evaluated at neighboring strides (\emph{e.g.}, 6, 7, and 11), indicating that GDSNet can compensate for moderate sampling mismatch through its continuous geometric parameterization. Although stride 12 yields slightly lower accuracy, it still \XP{stably} achieves an MAE of 77.23 under matched-stride evaluation, highlighting the strong tolerance of the proposed representation to coarser discretization.

\textbf{Efficiency.} The sampling stride $s$ controls the allocation budget. For an $H\times W$ image, the number of sampled control points $V=|\mathcal{V}|$ satisfies $V\leq\lceil H/s\rceil\lceil W/s\rceil$. A planar triangulation with $h$ convex-hull vertices contains $N=2V-h-2\leq2V-5$ triangular faces, each associated with one Gaussian primitive. Thus, increasing $s$ reduces the upper bound on Gaussian allocation, while the actual allocation remains image-adaptive. Delaunay construction takes $\mathcal{O}(V\log V)$ time, and feature aggregation over the bounded-degree simplex dual graph takes $\mathcal{O}(N)$, yielding \textbf{quasi-linear} $\mathcal{O}(N\log N)$ complexity for topology construction and primitive processing. Detailed runtime, memory, and scaling measurements, including the dominant cost of topology construction, are provided in \emph{Suppl.~\ref{sec:supp_complex}}.

%% file: sec/5.tex
\section{Conclusion}
\label{sec:conclusion}

In this paper, we propose GDSNet, which formulates crowd counting as a continuous density reconstruction problem using Gaussian basis functions, moving beyond traditional grid- and point-based representations. To address the ill-posed nature of unconstrained continuous modeling, we introduce a control-point-guided parameterization that imposes structured geometric constraints on Gaussian primitives. This design enables stable learning while preserving spatial continuity and flexibility. Extensive experiments on challenging benchmarks demonstrate consistent improvements, with strong robustness to variations in test-time sampling density.
How to further accelerate the Delaunay triangulation will be a future direction.

\section{Acknowledgement}
This work was funded in part by the National Key R\&D Program of China (2025YFC3811300), the National Natural Science Foundation of China (62376070, 62606119, 62076195), the China Postdoctoral Science Foundation under Grant (2026M794773), and the Fundamental Research Funds for the Central Universities (AUGA5710011522).

%% file: sec/Appendix.tex
\newpage 
\appendix 

\vspace*{1cm}
{\huge\centering Supplementary Material for Gaussian Density Splatting Network\par}
\vspace{0.5cm}

\section{Discussion on Mathematical Presentation of Different Counting Paradigms}
\label{sec:suppl_representation}

Given an input image $I \in \mathbb{R}^{H \times W \times 3}$ and point annotations $Y=\{y_m\}_{m=1}^{M}$, existing methods estimate the crowd distribution via grid-based density regression or point-based detection. From a functional approximation perspective, different counting paradigms can be unified as estimating $\mathcal{D}(x)$ over the continuous image domain $\Omega = \mathbb{R}^2$ via a linear combination of specific \textbf{basis functions} with varying \textbf{spatial supports}:
\begin{equation}
    \hat{\mathcal{D}}(x) = \sum\nolimits_{i} \alpha_i \Psi_i(x; \xi_i),
\end{equation}
where $\Psi_i$ denotes the basis function whose spatial support is parameterized by $\xi_i$, $w_i \ge 0$ is the discrete magnitude, and the global integral yields the total count $\hat{M} = \int_{\Omega} \mathcal{D}(x)\,dx$.
In the following, we formalize grid-based and point-based methods under this unified view, and subsequently introduce GDSNet as a continuous mathematical approximation representation.

\textbf{grid-based density estimation.}
grid-based methods~\cite{zhang2016single, ma2019bayesian} discretize the image plane into a regular grid $\mathcal{G}_s$, where $s$ denotes the output stride, and predict a scalar density value for each grid cell. The corresponding continuous density can be viewed as a piecewise-constant approximation:
\begin{equation}
    \hat{\mathcal{D}}_{\mathrm{map}}(x)
    =
    \sum\nolimits_{u \in \mathcal{G}_s}
    \hat{d}_u \Pi_{\Omega_u}(x),
    \qquad
    \Pi_{\Omega_u}(x)=
    \begin{cases}
      1, & \|x-x_u\|_\infty \leq s/2,\\
      0, & \mathrm{otherwise}.
    \end{cases}
\end{equation}
Here $\hat{d}_u$ is the scalar density predicted for the $u$-th cell centered at $x_u$, and $\Pi_{\Omega_u}$ is a \textbf{zero-order box basis with a rigid grid-aligned support}. 
Under our unified framework, the magnitude $\alpha_i$ corresponds to the predicted scalar density $\hat{d}_u$ for the $u$-th grid cell. The basis $\Psi_i$ is instantiated as a zero-order box function $\Pi_{\Omega_u}(x)$. The spatial support parameter $\xi_i$ is strictly locked to the rigid grid with fixed cell size $s$.
While simple and effective, its spatial support is tightly coupled with the network architecture, and thus rigidly locked to the grid cell size prescribed by the downsampling stride. This rigidity severely limits representational flexibility, barring handling extreme sub-pixel congestions and arbitrary-resolution density sampling.

\textbf{Point-based prediction.}
Point-based methods regress on a discrete set of candidate proposals, predicting a confidence score $p_i$ and a coordinate $\mu_i$ for each proposal~\cite{song2021P2P, Lin_2025_p2r}. 
After confidence thresholding, the crowd distribution degenerates into a sparse measure supported on isolated points:
\begin{equation}
    \hat{\mathcal{D}}_{\mathrm{point}}(x)
    =
    \sum\nolimits_{i}
    \mathbb{I}(p_i>\eta)\delta(x-\mu_i),
\end{equation}
where $\mathbb{I}(\cdot)$ is the indicator function and $\delta(\cdot)$ is the \textbf{Dirac delta basis function}, which mathematically possesses \textbf{a zero-volume isolated support}. 
Under the unified framework, the magnitude $\alpha_i$ degenerates into a binary indicator $\mathbb{I}(p_i>\eta)$, and the basis $\Psi_i$ collapses into a Dirac delta function $\delta$. The support parameter $\xi_i$ comprises solely the isolated coordinate $\mu_i$.
Although point coordinates are continuous, the resulting density has zero-volume support and fails to explicitly model the spatial extent, anisotropy, or smooth occupancy of crowded regions.

\textbf{GDSNet: continuous Gaussian basis representation.}
To facilitate a representation that is both geometrically explicit and spatially continuous, we introduce the \textbf{exponential basis function} and formulate the density distribution as a superposition of 2D Gaussian primitives:
\begin{equation}
    \hat{\mathcal{D}}_{\mathrm{gds}}(x)
    =
    \sum\nolimits_{i}
    \rho_i \mathcal{N}(x;\mu_i,\Sigma_i),
\end{equation}
where $\mathcal{N}$ represents the bivariate Gaussian basis function, parameterized by a learnable continuous center $\mu_i$, an anisotropic covariance matrix $\Sigma_i$, and a non-negative density mass scalar $\rho_i$. 
Here, the magnitude $w_i$ is represented by a learnable non-negative density mass $\rho_i$ and the basis $\phi_i$ is an exponential Gaussian function. The spatial support parameter $\theta_i=\{\mu_i, \Sigma_i\}$ explicitly models both the continuous center $\mu_i$ and the anisotropic covariance matrix $\Sigma_i$.
Compared with box or Dirac bases, Gaussian bases provide smooth spatial support, explicit geometry, and analytical mass conservation:
\begin{equation}
    \hat{M}
    =
    \int_{\Omega}\hat{\mathcal{D}}_{\mathrm{gds}}(x)\,dx
    =
    \int_{\Omega} \sum\nolimits_{i}
    \rho_i \mathcal{N}(x;\mu_i,\Sigma_i) \,dx
    =  \sum\nolimits_{i}
    \rho_i \int_{\Omega} \mathcal{N}(x;\mu_i,\Sigma_i) \,dx
    =
    \sum\nolimits_{i=1}^{N}\rho_i ,
\end{equation}
where the approximation becomes exact when the Gaussian mass outside the image boundary is negligible or handled by the rasterization domain.

\section{Implementation Details}
\label{sec:supp_implementation}

All experiments are conducted on one NVIDIA RTX 4090 GPU, implemented in PyTorch 2.0.1 with CUDA 11.8. 
We use VGG19~\cite{simonyan2014vgg} with ImageNet-pretrained weights as the backbone. Gaussian Splatting is implemented using gsplat~\cite{ye2024gsplatopensourcelibrarygaussian}. The sampling stride for control point allocation is set to $s=10$, the loss balancing temperatures are $\lambda_1=\lambda_2=1$, and the shape regularization threshold is $\gamma=2$. 
The structural prior is generated by an off-the-shelf, frozen structural estimator from~\cite{zeng2025instantGI}, without additional training or task-specific fine-tuning, as detailed in~\emph{Suppl.~\ref{instantgi}}.
The reference ellipse fitted on each triangle simplex is obtained by least-squares fitting to the six points consisting of the three vertices and the three edge midpoints~\cite{fitzgibbon1996buyer}.
The Adam optimizer (batch size = 1 and learning rate = $10^{-5}$) is used with image crop sizes adjusted per dataset: 256×256 for ShanghaiTech, 384×384 for NWPU, and 512×512 for the remaining benchmarks. 

The pseudo code of model training and inference is provided in Algorithm~\ref{alg:gdsnet}. \CR{Rasterization is required during training for spatial density supervision. At inference, count-only evaluation returns the analytical mass sum, while density-map evaluation additionally renders the learned Gaussian field using its geometry and mass parameters.}

\begin{algorithm}[p]
\caption{Training and Inference Pipeline of GDSNet}
\label{alg:gdsnet}
\SetAlgoLined
\KwIn{Input image $I$, sampling stride $s$, annotations $Y$ (train only)\CR{, density-map output flag \texttt{return\_density}}.}
\KwOut{Updated parameters $\Theta$ (train), estimated count $\hat{M}$ (test), rendered map $\hat{\mathcal{D}}$ (optional for test).}

\tcc{Phase 1: Structural Simplexity-Guided Control-Point Allocation (Sec.~\ref{subsec:point_allocation})}
Extract dense semantic feature map $\mathcal{F} \leftarrow \text{Backbone}_{\Theta}(I)$\;
Generate structural prior map $\mathbf{M}^e \leftarrow \text{Prior}(I)$\;
Apply max-pooling (stride $s$) and dithering~\cite{franchini2019stochastic} to $\mathbf{M}^e$ to yield binary activations $z_{i,j}$\;
Allocate discrete control points $\mathcal{V} \leftarrow \left\{ (is+\tfrac{s}{2}, js+\tfrac{s}{2}) \mid z_{i,j}=1 \right\}$\;
Construct simplicial mesh $\mathcal{T}$ and topological graph $\mathcal{G}(\mathcal{T}, \mathcal{E}) \leftarrow \text{DelaunayTriangulation}(\mathcal{V})$\;

\BlankLine
\tcc{Phase 2: Topology-Aware Feature Aggregation (Sec.~\ref{subsec:parameterization})}
\ForEach{simplex $\tau_i \in \mathcal{T}$}{
    Fit base reference ellipse~\cite{fitzgibbon1996buyer} $e_i$ and extract its geometric priors $\{r_i^0, \theta_i^0\}$\;
    Construct \textbf{geometric feature} $f_i^g$ using normalized vertex coordinates and $e_i$\;
    Sample $\mathcal{F}$ at center $c_i$ and vertices $v_{i,j}$ to extract \textbf{appearance feature} $f_i^a$ (Eq.~\ref{eq:app_feature})\;
    Fuse intra-simplex representation $f_i^l \leftarrow f_i^g \oplus f_i^a$\;
Aggregate graph-induced contextual features from neighbors $b \in \mathcal{B}(i)$ to form $f_i^t$ (Eq.~\ref{eq:fi})\;
    Concatenate to form the comprehensive primitive feature: $f_i \leftarrow f_i^l \oplus f_i^t$\;
}

\BlankLine
\tcc{Phase 3: Constrained Parameterization (Sec.~\ref{subsec:parameterization})}
\ForEach{feature $f_i$}{
    Regress relative adjustments and density mass via parallel MLPs: 
    $\mathbf{w}_i, \Delta\mathbf{s}_i, \Delta\theta_i, \rho_i \leftarrow \text{MLPs}_{\Theta}(f_i)$\;
    Explicitly restrict center $\mu_i$ within the simplex convex hull via $\mathbf{w}_i$: $\mu_i \leftarrow \sum\nolimits_{j=1}^3 w_{i,j} v_{i,j}$ (Eq.~\ref{eq:mu_sigma})\;
    Map scale and rotation relative to base priors: $\mathbf{s}_i \leftarrow \Delta\mathbf{s}_i \cdot [r_i^0, 1]^\top$, $\theta_i \leftarrow \theta_i^0 + \Delta \theta_i$ (Eq.~\ref{eq:mu_sigma})\;
    Instantiate continuous Gaussian primitive: $\mathcal{G}_i \leftarrow \{\mu_i, \mathbf{s}_i, \theta_i, \rho_i\}$\;
}

\BlankLine
\tcc{Phase 4: State Execution \& Optimization (Sec.~\ref{subsec:optimization})}
Compute analytical total count by summing the density mass: $\hat{M} \leftarrow \sum_{i=1}^{|\mathcal{T}|} \rho_i$\;
\eIf{is Training Stage}{
    Rasterize primitives $\{\mathcal{G}_i\}$ to continuous density map $\hat{\mathcal{D}}$~\ref{eq:gsc_density} via Gaussian splatting\;
    Compute spatial alignment Bayesian loss $\mathcal{L}_{\mathrm{rast}}(\hat{\mathcal{D}}, Y)$ (Eq.~\ref{eq:rast_loss})\;
    Compute global count loss $\mathcal{L}_{\mathrm{cnt}}$ (Eq.~\ref{eq:cnt_loss}) and geometric shape loss $\mathcal{L}_{s}$ (Eq.~\ref{eq:shape_loss})\;
    Optimize overall objective $\mathcal{L} = \mathcal{L}_{\mathrm{rast}} + \lambda_1\mathcal{L}_{\mathrm{cnt}} + \lambda_2\mathcal{L}_{s}$ (Eq.~\ref{eq:overall}) via backpropagation to update $\Theta$\;
}{
    
    \eIf{\texttt{return\_density}}{
        Rasterize $\{\mathcal{G}_i\}$ to obtain $\hat{\mathcal{D}}$\;
        \Return $\hat{M},\hat{\mathcal{D}}$\;
    }{
        \Return $\hat{M}$\;
    }
    
}
\end{algorithm}

\section{Complexity and Scalability Analysis}
\label{sec:supp_complex}

A critical concern for crowd counting in extremely dense scenarios is the scalability of the representation. 
We analyze the theoretical complexity of GDSNet with respect to the number of allocated Gaussians $N$, which corresponds to the number of triangular simplices. 
Unlike transformer-based architectures that typically suffer from quadratic $\mathcal{O}(N^2)$ or $\mathcal{O}(H^2W^2)$ complexity due to global self-attention, GDSNet maintains a highly efficient scalability profile. 

The process begins with the Delaunay triangulation of $V$ control points, which operates in quasi-linear $\mathcal{O}(V \log V)$ time. 
Since the number of triangular faces $N$ in a planar triangulation is linearly proportional to the number of vertices ($N \approx 2V$), this stage effectively scales as $\mathcal{O}(N \log N)$. 
The subsequent \textit{Topology-Aware Aggregation} is performed over the dual graph of the triangulation, where each Gaussian (simplex) acts as a node. 
In such a graph, every node is adjacent to at most three neighbors. This \textbf{bounded degree} ensures that the total number of edges in the topological graph is strictly linear, $\mathcal{O}(N)$. 
Consequently, the message-passing complexity for feature aggregation remains strictly linear, $\mathcal{O}(N)$.

Furthermore, the memory footprint of our continuous representation scales linearly with $N$ ($\mathcal{O}(N)$), contrasting favorably with dense volumetric or high-resolution grid representations. 
During the rasterization phase, tile-based rendering is bounded by the image resolution $\mathcal{O}(H \times W)$ and is completely \textbf{decoupled} from the crowd density. 
This $\mathcal{O}(N \log N)$ overall bound ensures that GDSNet scales robustly even on challenging datasets like JHU-CROWD++ containing tens of thousands of individuals, without encountering out-of-memory (OOM) failures or exponential inference delays.

\subsection{Component-wise Efficiency}

Since GDSNet contains multiple components, we provide a detailed component-wise analysis to identify the computational cost distribution of each module. Table~\ref{tab:cr_efficiency} reports the parameter, GPU memory, and runtime costs of the main components when processing a $256\times256$ input image. Values prefixed by ``+'' denote the additional cost of each component.

\begin{table}[htbp]
\centering\small
\caption{Component-wise parameter, GPU memory, and runtime measurements.}
\label{tab:cr_efficiency}
\begin{tabular}{lrrr}
\toprule
Component & Params (M) & GPU memory (MB) & Runtime (ms)\\
\midrule
Baseline: backbone + regression head & 21.50 & 130.50 & 12.54\\
Backbone & 20.02 & 48.64 & 11.02\\
+ Topology computation & +0 & +0.65 & +9.00\\
+ Feature aggregation & +1.66 & +11.04 & +0.29\\
+ Constrained Gaussian parameterization & +0.04 & +1.70 & +0.20\\
+ Gaussian field rasterization & +0 & +0.21 & +0.11\\
\bottomrule
\end{tabular}
\end{table}

The analysis shows that the Gaussian field learning and rendering modules themselves introduce only marginal computational overhead. 
The feature aggregation, constrained Gaussian parameterization, and Gaussian field rasterization together require only 0.60 ms and 12.95 MB memory, demonstrating that the Gaussian parameter learning and differentiable rendering modules are lightweight.

The main additional overhead comes from the topology computation. This overhead is mainly caused by the Delaunay triangulation step that explicitly constructs spatial relationships among adaptive control points. Importantly, this operation is non-parametric and does not introduce additional trainable parameters. 

In future work, more advanced GPU-accelerated triangulation tools or learned primitive organization strategies can be investigated to further reduce the overhead of topology construction. 

\subsection{Scaling with Primitive Allocation}
We vary the sampling stride on the same input image under an identical inference setting and exclude image loading time. Table~\ref{tab:cr_scaling} separately reports topology construction to isolate the principal primitive-dependent cost. Smaller strides allocate more Gaussians and increase topology construction time. GPU memory varies little over the measured range.

\begin{table}[!htbp]
\centering
\caption{Runtime and GPU memory measurements with varying Gaussian primitive allocation.}
\label{tab:cr_scaling}
\resizebox{\linewidth}{!}{\begin{tabular}{rrrrrr}
\toprule
Stride & Gaussian & Runtime (ms) & GPU memory (MB) & Topology time (ms) & Topology memory (MB)\\
\midrule
6 & 1814 & 16.93 & 48.95 & 14.37 & 1.06\\
8 & 1002 & 13.65 & 48.93 & 11.15 & 0.56\\
10 & 639 & 12.27 & 48.92 & 9.72 & 0.55\\
12 & 451 & 10.94 & 48.92 & 8.42 & 0.55\\
16 & 244 & 10.07 & 48.91 & 7.59 & 0.51\\
\bottomrule
\end{tabular}}
\end{table}
\FloatBarrier

The results show that the number of allocated Gaussian primitives can be flexibly adjusted through the stride $s$. The topology computation dominates the primitive-dependent scaling cost and increases consistently as more simplices are allocated. Importantly, as the number of simplices/Gaussians increases, the topology time and overall runtime scale approximately linearly, which empirically validates our claim of quasi-linear complexity. The other costs remain nearly unchanged. These results verify that the computational overhead of the proposed Gaussian representation is controllable with respect to the number of allocated primitives.

\section{Implementation Details of Ablation Studies}

To investigate the modules in the proposed GDSNet, we implement module ablation and Control Point Allocation~(CPA) strategy ablation studies, as shown in Table~\ref{tab:supp_module_comparison}.

\begin{table}[ht]
\centering
\caption{Implementation comparison of the full model and ablation variants. CPA, SCP, and TFA are the control point allocation, simplex-constrained parameterization, and topology-aware feature aggregation modules, respectively.}
\label{tab:supp_module_comparison}
\resizebox{\linewidth}{!}{
\begin{tabular}{l|lccccc}
\toprule
& \textbf{Method} 
& \textbf{CPA Strategy} 
& \textbf{Sampling} 
& \textbf{Support Region} 
& \textbf{SCP} 
& \textbf{TFA} \\
\midrule

\multirow{4}{*}{Module Ablations}
& Full GDSNet 
& Structural Prior
& Dither 
& Triangle Simplex
& \checkmark 
& \checkmark \\

& w/o TFA
& Structural Prior
& Dither 
& Triangle Simplex 
& \checkmark 
& -- \\


& w/o (TFA \& SCP)
& Structural Prior 
& Dither 
& Fixed grid patch 
& -- 
& -- \\

& w/o (TFA \& SCP \& CPA  )
& Structural Prior  
& Uniform grid 
& Fixed grid patch 
& -- 
& -- \\

\hline
\multirow{3}{*}{\shortstack{CPA\\ Ablation}}
& Uniform Grid
& Uniform 
& Dither 
& Triangle Simplex
& \checkmark 
& \checkmark \\

& Sobel Filter Response Prior
& Sobel Filter Response Prior
& Dither 
& Triangle Simplex
& \checkmark 
& \checkmark \\

& Structural Prior
& Structural Prior 
& Dither 
& Triangle Simplex
& \checkmark 
& \checkmark \\
\bottomrule
\end{tabular}
}
\end{table}

\subsection{Implementation of Module Ablations}
\label{sec:supp_module_ab}

\paragraph{Removing Topology-aware Feature Aggregation (TFA).}
This variant removes the inter-simplex neighbor feature while keeping control point allocation, dither sampling, and simplex-constrained parameterization. The final feature for Gaussian becomes:
\begin{equation}
    f_i=f_i^{l},
\end{equation}
instead of concatenating the context feature of one-hop Delaunay neighbors. This setting evaluates the contribution of local topological context to scale, rotation, and density mass prediction.

\paragraph{Removing Topology-aware Feature Aggregation (TFA) and Simplex-Constrained Parameterization (SCP).}
This variant also removes neighbor modeling of the Gaussian feature, and replaces the adaptive triangular receptive field with a fixed square grid patch. For each sampled anchor $v_i$, four patch corners are defined as:
\begin{equation}
    q_{i,1}=v_i+(-\Delta_x,-\Delta_y),\quad
    q_{i,2}=v_i+(+\Delta_x,-\Delta_y),
\end{equation}
\begin{equation}
    q_{i,3}=v_i+(-\Delta_x,+\Delta_y),\quad
    q_{i,4}=v_i+(+\Delta_x,+\Delta_y),
\end{equation}
where $\Delta_x=s/W$ and $\Delta_y=s/H$ in normalized coordinates. The local feature is obtained by sampling the backbone feature map at the four patch corners and the center
\begin{equation}
    f_i^{l}
    =
    \mathcal{F}(q_{i,1})
    \oplus
    \mathcal{F}(q_{i,2})
    \oplus
    \mathcal{F}(q_{i,3})
    \oplus
    \mathcal{F}(q_{i,4})
    \oplus
    \mathcal{F}(v_{i}).
\label{eq:w/o_tfa}
\end{equation}

Without the constraint of the triangle simplex, this variant feeds the Gaussian feature into MLP heads and gets $ \{ \Delta \mu,\Delta \mathbf{s}_i, \Delta \theta_i, \rho_i \}$. The parameters of the Gaussian primitive are:
\begin{equation}
\mu_i = v_i + \Delta \mu_i, \quad \mathbf{s}_i = \Delta \mathbf{s}_i \cdot \begin{bmatrix} 1 , 1 \end{bmatrix}^\mathsf{T}, \quad \theta_i = 0 + \Delta \theta_i, \quad \rho_i = \rho_i.
\label{eq:w/o_scp}
\end{equation}

\paragraph{Removing Topology-aware Feature Aggregation (TFA), Simplex-Constrained Parameterization (SCP), and Structural Complexity Guided Control Point Allocation (CPA).}
This is the most reduced variant. It removes the structural prior and instead places Gaussian anchors uniformly on the image plane:
\begin{equation}
    \mathcal{V}_{\mathrm{grid}}
    =
    \{(k/2+ik,\;k/2+ik)\}_{i,j}.
\end{equation}
Each anchor uses the same fixed grid-patch receptive field as described above. The feature extraction and Gaussian parameterization follows Eq.~\ref{eq:w/o_tfa} and Eq.~\ref{eq:w/o_scp}. This experiment evaluates whether uniform spatial allocation with local patch features is sufficient for Gaussian density rendering.

\subsection{Implementation and Visualization of Control Point Allocation Strategy Ablations}
\label{sec:supp_CPA_ab}

\paragraph{Uniform Grid Allocation.}
In this setting, the structural prior map is replaced by a constant map:
\begin{equation}
    \mathbf{M}^{e}(x)=1,\quad \forall x\in\Omega.
\end{equation}
The remaining pipeline is unchanged. This variant isolates the influence of learned complexity guidance from the rest of the topology-aware Gaussian feature aggregation pipeline.

\paragraph{Sobel Filter Response Prior.}
This variant constructs a prior map from handcrafted image structural prior as in Image-gs~\cite{zhang2025imagegs}. The input image is first converted to grayscale:
\begin{equation}
    Y=0.299R+0.587G+0.114B.
\end{equation}
Then, Sobel filters are applied:
\begin{equation}
    G_x=K_x*Y,\qquad G_y=K_y*Y,
\end{equation}
and the response magnitude is computed as:
\begin{equation}
    G=\sqrt{G_x^2+G_y^2}.
\end{equation}
After max normalization, the response is squared to emphasize strong edges:
\begin{equation}
    \mathbf{M}^{e}_{\mathrm{grad}}
    =
    \left(
    \frac{G}{\max(G)+\epsilon}
    \right)^2.
\end{equation}
To avoid overly sparse sampling in flat regions, a small lower bound is applied:
\begin{equation}
    \mathbf{M}^{e}_{\mathrm{grad}}
    =
    \max(\mathbf{M}^{e}_{\mathrm{grad}},\rho),
\end{equation}
where $\rho$ is a small constant. The subsequent pipeline is identical to the full model.

\paragraph{Structural Complexity Prior.}
\label{instantgi}
This is the default setting of GDSNet. The prior map is generated by the off-the-shelf, frozen structural estimator from in instantGI~\cite{zeng2025instantGI}.
Benefiting from its exceptional generalization across diverse scenes, we \textit{directly employ this plug-and-play component to generate $\mathbf{M}^{e}$ for all experiments}. It is important to note that the InstantGI model is used \textit{purely for inference in our pipeline to produce the structural prior}, and is \textit{not fine-tuned or trained} on any target dataset. By utilizing this task-agnostic substrate, GDSNet can focus exclusively on task-specific density regression, effectively avoiding optimization instabilities while ensuring robust performance across varied crowd counting benchmarks.

As for the architecture of this estimator, $\mathbf{M}^{e}$ is predicted by a ConvNeXt-UNet followed by a pixel-wise MLP:
\begin{equation}
    \mathbf{F}_{e}=f_{\mathrm{SPM}}(I), 
    \qquad 
    \mathbf{M}^{e}=\sigma(\phi(\mathbf{F}_{e})),
\end{equation}
where $\mathbf{F}_{e}\in\mathbb{R}^{64\times H\times W}$ and $\phi$ is a three-layer MLP with channel dimensions $64\rightarrow64\rightarrow64\rightarrow1$. $\sigma$ is the Sigmoid activation function.

\begin{figure}
    \centering
    \includegraphics[width=1.0\linewidth]{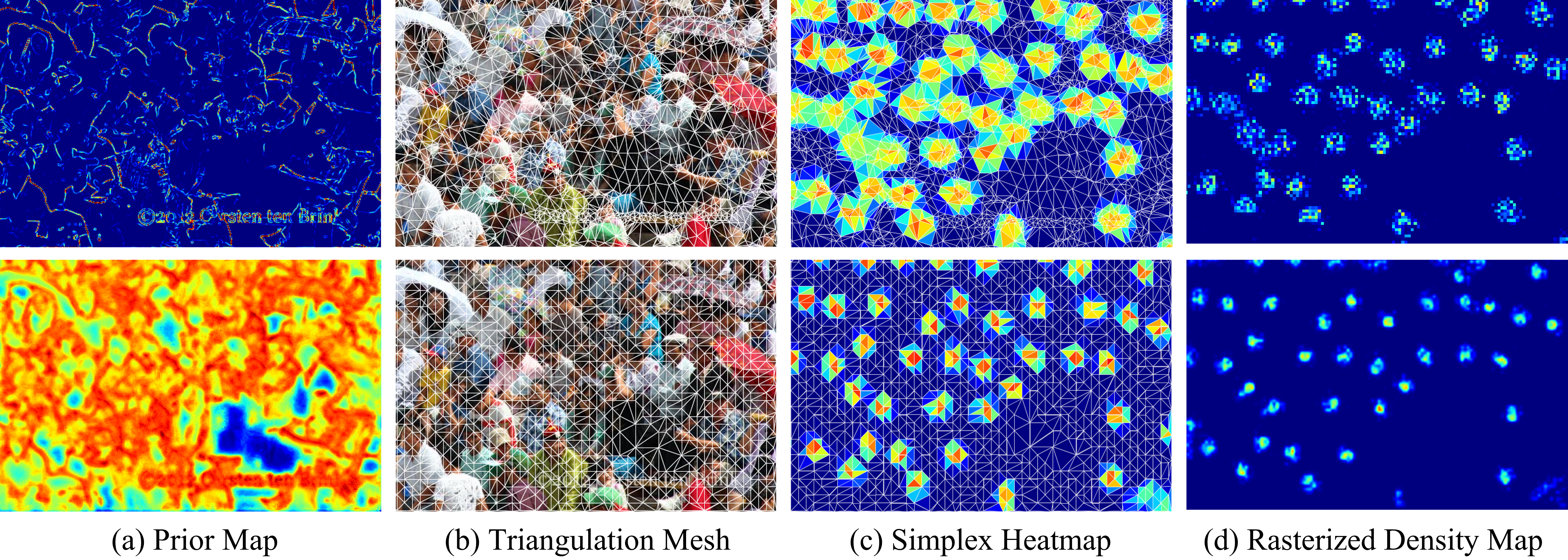}
    \caption{Comparison of Sobel filter response prior (top) and our structural prior (bottom).}
    \label{fig:supp_prior}
\end{figure}

\textbf{More Visual Comparison of CPA Strategies.} To further elucidate the impact of different priors, we visualize the intermediate stages of GDSNet using Sobel filtering versus our structural prior in Fig.~\ref{fig:supp_prior}. As shown in the top row, the Sobel-generated map is strictly contour-focused, capturing high-frequency edges and irrelevant background noise (e.g., the watermark at the bottom). This leads to an overly fragmented mesh that forces the Gaussian primitives to collapse into thin, erratic shapes, resulting in the noisy and disjointed density field. In contrast, our structural complexity-guided prior (bottom row) identifies geometric manifolds and salient clusters rather than just edges. This provides a more balanced triangulation substrate, allowing the network to regress smooth, ellipsoid-like primitives that accurately represent the spatial occupancy of individuals. This qualitative evidence confirms that our choice of structural prior is essential for maintaining the topological continuity of the crowd density field. Ultimately, this stable initialization effectively regularizes the learning process by anchoring the primitive topology, enabling the model to prioritize precise density mass estimation over topological searching. This synergy translates observed visual coherence into superior counting accuracy.

\subsection{Implementation Details of Gaussian Geometry Ablation}
\label{sec:cr_geometry}
All variants share the same control points, primitive allocation, feature aggregation module, density mass prediction head, and rasterization supervision. They differ in the spatial basis and its geometric parameterization.

\textbf{Full anisotropic Gaussian parameterization.} After Delaunay triangulation, a reference ellipse $e_i$ is fitted to each simplex $\tau_i$ to establish its geometric prior. From the aggregated feature representation, the network predicts barycentric weights $w_{i,j}$, scale adjustments $\Delta\mathbf{s}_i$, a rotation adjustment $\Delta\theta_i$, and density mass $\rho_i$. Following Eq.~\ref{eq:mu_sigma}, the final geometry is
\begin{equation}
\mu_i=\sum_{j=1}^{3}w_{i,j}v_{i,j},\qquad
\mathbf{s}_i=\Delta\mathbf{s}_i\odot[r_i^0,1]^\mathsf{T},\qquad
\theta_i=\theta_i^0+\Delta\theta_i,
\end{equation}
where $r_i^0$ and $\theta_i^0$ are the axis ratio and orientation of the fitted ellipse. The barycentric weights constrain the center to its simplex. The following variants modify this formulation while retaining the shared mass prediction and supervision.

\textbf{Triangle mass regression.} This variant uses a unit-integral, piecewise-constant triangle basis:
\begin{equation}
\Psi_i(x)=\frac{\mathbf{1}[x\in\tau_i]}{|\tau_i|},\qquad
\hat{\mathcal{D}}(x)=\sum_i\rho_i\Psi_i(x),
\end{equation}
where $|\tau_i|$ is the triangle area. The predicted mass is uniformly distributed over each triangle.

\textbf{Fixed Gaussian.} This variant removes learnable geometric adjustments: $\mu_i$, $\mathbf{s}_i$, and $\theta_i$ are directly set to the center, axis scales, and orientation of the fitted ellipse $e_i$ and remain fixed during training. The density mass $\rho_i$ is still predicted by the mass head.

\textbf{Isotropic Gaussian.} This variant removes anisotropic aspect-ratio modeling by setting $r_i^0=1$ and tying the two scale dimensions:
\begin{equation}
\mathbf{s}_i=\Delta s_i[1,1]^\mathsf{T}.
\end{equation}
The center retains the full model's learnable barycentric parameterization, and $\Delta s_i$ is a learnable scalar scale adjustment. Thus, the variant preserves adaptive center and scale estimation while removing directional anisotropy. Rotation does not affect an isotropic Gaussian's spatial support.

These variants provide a progressive comparison from a piecewise-constant triangle basis to fixed Gaussian support, learnable isotropic support, and full anisotropic geometry. The quantitative results and analysis are provided in Table~\ref{tab:cr_geometry} in the main paper.

\section{Comparison with MLLMs}
\label{sec:cr_mllm}
MLLMs and specialized crowd counting networks leverage distinct strengths and cater to different application paradigms.

MLLMs excel at high-level semantic understanding, open-world reasoning, and object-centric counting in sparse scenarios with distinct instances, such as the FSC-147 benchmark which averages approximately 56 objects per image. Specialized networks like GDSNet are purpose-built for ultra-dense human scenes characterized by extreme scale variations, severe occlusion, and perspective distortion, where benchmarks like ShanghaiTech Part A average 501 persons per image and UCF-QNRF averages 815 persons per image.

Directly applying MLLMs to dense crowd benchmarks remains highly challenging due to visual tokenization resolution limits and auto-regressive numeric regression bottlenecks. Table~\ref{tab:cr_mllm} compares reported ShanghaiTech Part A results, highlighting this performance gap. External model results are taken from CountGD++~\cite{AminiNaieni26b} and WS-COC~\cite{zhang2026bootstrapping}.

\begin{table}[htbp]
\centering\small
\caption{Reported MLLM and GDSNet results on ShanghaiTech Part A.}
\label{tab:cr_mllm}
\begin{tabular}{llr}
\toprule
Model / source & Setting & MAE\\
\midrule
Molmo~\cite{AminiNaieni26b} & Direct prompting & $7.26\times10^8$\\
Gemini-2.5~\cite{AminiNaieni26b} & Direct prompting & 517.17\\
LLaVA-OneVision-7B~\cite{zhang2026bootstrapping} & Direct prompting & 336.80\\
LLaVA-OneVision-7B / WS-COC~\cite{zhang2026bootstrapping} & Counting-specific tuning & 128.90\\
GDSNet & Specialized density estimation & 48.80\\
\bottomrule
\end{tabular}
\end{table}
\FloatBarrier

Table~\ref{tab:cr_mllm} indicate a substantial gap between general-purpose MLLMs and specialized models on dense crowd counting. Visual tokenization can limit access to small, heavily occluded heads, while numerical count prediction introduces an additional challenge. Counting-specific tuning improves MLLM performance, but specialized density estimation remains valuable in these dense scenes.

\section{Failure Cases and Limitations}
\label{sec:cr_failure}
\begin{figure}[htbp]
    \centering
    \includegraphics[width=1.0\linewidth]{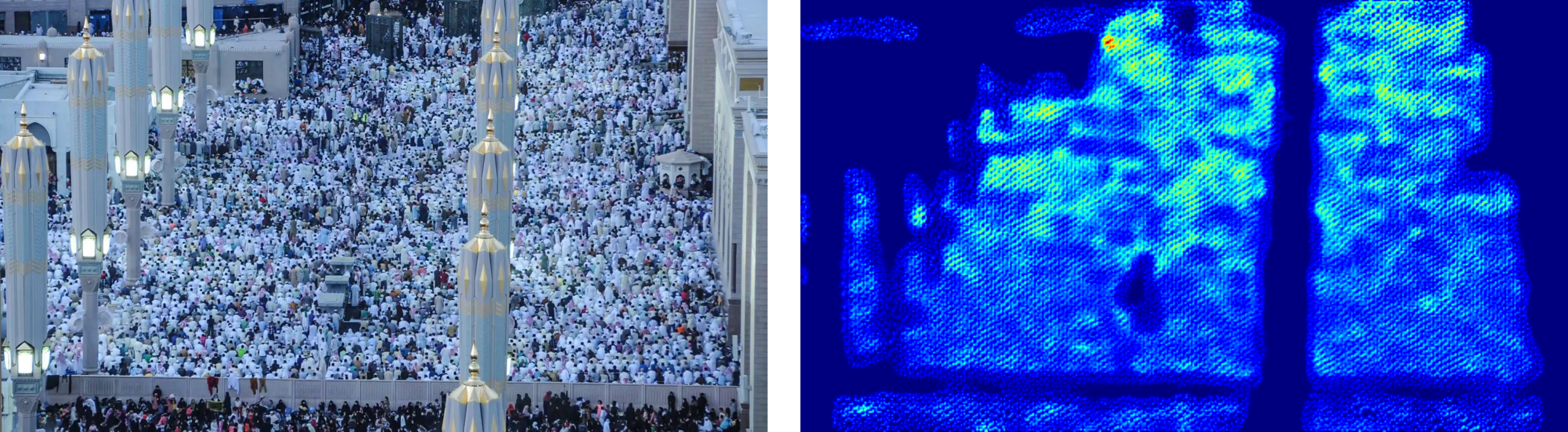}
    \caption{Representative failure case of GDSNet on a highly congested scene. The model produces a smooth density map which lacks sharply localized peaks for individual heads.}
    \label{fig:supp_failure}
\end{figure}

In extremely congested scenes, individual heads may occupy only a few pixels. GDSNet can produce a reasonably accurate total count while rendering smooth high-response regions rather than sharply localized peaks, as shown in Fig.~\ref{fig:supp_failure}. Constraining primitives against excessively small support improves training stability but limits response sharpness. The current supervision emphasizes density alignment and count consistency without a strong individual-head localization constraint.
Two directions for improvement are extracting centers of high-response primitives or sharpening responses at inference, and jointly optimizing a localization-oriented objective to learn sharper spatial responses alongside accurate counting.

